\documentclass{IEEEtran}
\usepackage{setspace}
\usepackage[T1]{fontenc}
\usepackage[latin9]{inputenc}
\usepackage{color}
\usepackage{verbatim}
\usepackage{float}
\usepackage{units}
\usepackage{mathtools}
\usepackage{amsmath}
\usepackage{amssymb}
\usepackage{graphicx}
\usepackage{setspace}

\makeatletter

\floatstyle{ruled}
\newfloat{algorithm}{tbp}{loa}
\providecommand{\algorithmname}{Algorithm}
\floatname{algorithm}{\protect\algorithmname}

\usepackage{eqnarray}
\usepackage{mathrsfs}
\usepackage{epsfig}
\usepackage[english]{babel}
\usepackage{subfigure}
\usepackage{caption}
\usepackage{epstopdf}
\usepackage{import}

\usepackage{colortbl}\usepackage{cite}
\usepackage{algorithm}

\usepackage{bbm}
\usepackage{cases}
\usepackage[normalem]{ulem}
\usepackage{array}
\usepackage{times}
\usepackage{amsfonts}
\usepackage{psfrag}
\usepackage{fancyhdr}
\usepackage{algorithmic}
\allowdisplaybreaks

\usepackage{soul}

\usepackage{dsfont}

\usepackage{babel}

\makeatother

\begin{document}
\title{Cooperative Multi-Agent Learning for Navigation via Structured State Abstraction\thanks{This work is funded by the European Union through the projects 6G-INTENSE (G.A no. 101139266), and CENTRIC (G.A no. 101096379). Views and opinions expressed are however those of the author(s) only and do not necessarily reflect those of the European Union. Neither the European Union nor the granting authority can be held responsible for them. This research was supported by the Research Council of Finland (former Academy of Finland) 6G Flagship Program (Grant Number: 346208)}}
\author{\IEEEauthorblockN{Mohamed~K.~Abdel-Aziz,~\IEEEmembership{Graduate Student Member,~IEEE}, Mohammed~S.~Elbamby,~\IEEEmembership{Member,~IEEE}, Sumudu~Samarakoon,~\IEEEmembership{Member,~IEEE},
and Mehdi~Bennis,~\IEEEmembership{Fellow,~IEEE}}\thanks{M.~K.~Abdel-Aziz, and M.~S.~Elbamby are with Nokia Bell Labs, Espoo, Finland
(e-mails: \{mohamed.3.abdelaziz, mohammed.elbamby\} @nokia-bell-labs.com).}\thanks{S.~Samarakoon, and M.~Bennis are with the Centre
for Wireless Communications, University of Oulu, Oulu, Finland
(e-mails: firstname.lastname@oulu.fi).}
 \thanks{M.~K.~Abdel-Aziz contributed to this work while being with the Centre
for Wireless Communications, University of Oulu.}}
\maketitle

\begin{abstract}
  Cooperative multi-agent reinforcement learning (MARL) for navigation enables agents to cooperate to achieve their navigation goals. Using emergent communication, agents learn a communication protocol to coordinate and share information that is needed to achieve their navigation tasks. In emergent communication, symbols with no pre-specified usage rules are exchanged, in which the meaning and syntax emerge through training. Learning a navigation policy along with a communication protocol in a MARL environment is highly complex due to the huge state space to be explored.  To cope with this complexity, this work proposes a novel neural network architecture, for jointly learning an adaptive state space abstraction and a communication protocol among agents participating in navigation tasks. The goal is to come up with an adaptive abstractor that significantly reduces the size of the state space to be explored, without degradation in the policy performance. Simulation results show that the proposed method reaches a better policy, in terms of achievable rewards, resulting in fewer training iterations compared to the case where raw states or fixed state abstraction are used. Moreover, it is shown that a communication protocol emerges during training which enables the agents to learn better policies within fewer training iterations.\newline
\end{abstract}


\section{Introduction\label{sec:Introduction}}

Shannon and Weaver categorized communication problems into three levels \cite{ShannonWeaver}: \emph{Level A} asks the question of how accurately can the communication symbols be transmitted (The technical problem), while \emph{Level B} asks the question of how precisely do the transmitted symbols convey the desired
meaning (The semantic problem), and finally \emph{Level C} asks the question of how effectively does the received meaning affect conduct in the
desired way (The effectiveness problem). Although the technical problem is important and has been studied extensively in the past, it is not the only factor that determines the success of a communication. The semantic and effectiveness problems are equally important and should be carefully considered when designing a communication system, especially for complex tasks such as field-of-view (FoV)-based navigation \cite{DRL_RobotNavigation}.

FoV-based navigation is a task where a group of vehicles (also referred to as agents) must navigate to an unknown destination using only their local observations and local sensory data, i.e., each agent\footnote{agent, vehicle, and robot are used interchangeably throughout the paper.} can only see a small portion of the environment around it. In order to coordinate their actions effectively, agents need to be able to communicate the meaning of their local observations to each other. This is the semantic problem. For example, an agent might need to communicate to others that it has seen the destination, or that its observed region is full of obstacles, etc. In addition, FoV-based navigation agents need to be able to act on the information that they receive from other agents. This is the effectiveness problem. For example, if an agent receives a message from another agent that it has seen the destination, it needs to be able to plan a new route to reach that destination. Hence, Levels B and C are critical for FoV-based navigation because agents need to be able to communicate the meaning of their observations and act on the information that they receive from other agents in order to cooperate and achieve their goal.


Traditionally, navigation tasks can be solved independently by each agent, by combining multiple modules, such as simultaneous localization and mapping (SLAM) \cite{ORB-SLAM,GMaping-SLAM}, path planning \cite{Sampling-BasedMotionPlanning}, and so forth. Integrating such modules in practical applications involves large computational errors, leading to poor performance \cite{DRL_RobotNavigation}. Moreover, a high-precision global map is required for such a traditional framework, resulting in limitations in navigating in unknown or dynamic environments.
One of the key challenges of FoV-based navigation is that agents need to be able to make navigation decisions effectively even though they can only see a small portion of the environment around them. This contrasts with traditional navigation approaches, which rely on a high-precision global map.

To address this challenge, and by utilizing the remarkable capabilities of deep learning technology,  deep reinforcement learning (DRL) has been used to directly learn navigation strategies from raw sensory inputs \cite{DRL_RobotNavigation,NavigationRL1,NavigationRL2}. In \cite{NavigationRL1}, a target-driven concept is proposed where the DRL algorithm is found to be generalizable to new scenarios, but at the expense of a decrease in performance. While in \cite{NavigationRL2}, a new neural network architecture was proposed in order to jointly learn a navigation task via RL along with additional auxiliary tasks. However, this approach was found to have an unstable policy and poor robustness in a complex environment \cite{DRL_RobotNavigation}.


%
Although DRL is shown to be a promising approach to FoV-based navigation, it has some limitations, including dependence on high-dimensional raw noisy local observations, decreased performance when generalizing to new scenarios, and lack of cooperation between agents to solve the navigation task. Real-world applications often require multiple agents to navigate together, which imposes extra challenges, such as the need for agents to cooperate and communicate effectively to achieve their goals. 
To tackle this, some works have utilized multi-agent reinforcement learning (MARL), to solve complex FoV-based navigation tasks \cite{MARLVision2,MARLVision3,MARLVision1}. In \cite{MARLVision2,MARLVision3}, the authors are interested in understanding how two agents can learn, from pixels, to communicate so as to effectively and collaboratively solve a given task. While in \cite{MARLVision1}, the authors study the cooperative navigation of multiple agents across photo-realistic environments to reach target locations, where each agent is equipped with a private external memory to persistently store communication information. Although interesting, the communicated messages are high dimensional \cite{MARLVision1}, MARL algorithms within these works still depends on high-dimensional visual input, and the agents share their policy parameters \cite{MARLVision2,MARLVision3}.

Agents can communicate by sharing their local observations, positions, goals, intentions, etc. However, communication channels between agents are often bandwidth-limited, so careful consideration must be given to which information to communicate, as this can impact the success of the task. The large number of states and information available to each agent makes it a daunting task to manually design such a communication protocol.
Moreover, the resulting protocol may not necessarily be optimal for the given task \cite{HoydisMACLearn}. Alternatively, agents could learn communication protocols to coordinate and share information that is needed to complete the tasks in a much faster and robust manner \cite{FoersterEC}. Learning a communication protocol in an MARL environment is referred to as emergent communication. In this setting, communication is emergent in the sense that, at the beginning of training, the symbols emitted by the agents have no predefined semantics nor pre-specified usage rules. Instead, the meaning and syntax emerge through training \cite{Emergent_Comm-Lazaridou}.
Emergent communication has been successfully applied to some navigation tasks \cite{LearnToCooperate,EffectiveComm}, where the authors have shown that agents are able to learn an interpretable, grounded communication protocol that allows them to efficiently solve navigation tasks in various grid world environments. However, both works considered simple navigation maps, either low dimensional \cite{LearnToCooperate}, or Obstacle-free \cite{EffectiveComm}.

Nonetheless, having multiple learning agents within the same environment poses an extra challenge to the learning problem as the problem's complexity increases exponentially with the number of agents \cite{MARLsurvey}. According to \cite{MARLsurvey,IsGoodRep}, DRL and MARL usually require millions of interactions for an agent to learn, due to the huge state space that needs to be explored to learn a suitable policy. Moreover, learning a communication protocol on top of that deteriorates the exploration problem because the state space grows even more with the added communication, and the likelihood for a consistent protocol to be found becomes difficult \cite{GroundEmergentComm}. 

To tackle the huge state space problem, state abstraction (compression/quantization) could be considered. State abstraction is a method for compressing the environment's state space to transform complex problems into simpler ones with more compact forms \cite{StateAbstractionApprenticeship}. Abstraction gives rise to models of the environment that are compressed and useful for downstream tasks, enabling efficient training, decision making and generalization. State abstraction could be carried out by eliminating unimportant states or combining states into groups to create a smaller state space. The state abstraction problem has been tackled in a number of previous works, i.e., \cite{Task-Oriented,StateAbstractionApprenticeship}. Due to the limited bandwidth communications between the agents, the authors of \cite{Task-Oriented} studied how agents should efficiently represent their local observations and communicate an abstracted version of it to improve the collaborative task performance. However, they assumed the presence of instantaneous and synchronous communications between agents. While in \cite{StateAbstractionApprenticeship}, the authors formulated and analysed the trade-off between abstraction and RL performance. Therein, the knowledge of the expert RL policy was assumed a priori. This expert policy is then used for learning  \emph{state abstractions}. Further, the size of the abstracted space was determined a priori. Finally, in order to allow agents to learn concepts from data, the state space needs to be structured \cite{StructureAbstraction}, so that agents can generalize better to new environments not seen during training.

\subsection{Contributions}

In this work, we propose a novel neural network architecture to jointly learn adaptive state space abstraction strategies, along with a communication protocol among agents. We focus on the semantic and effectiveness problems of communication with a specific emphasis on a FoV-based cooperative navigation scenario.\footnote{Assuming the perfect transmission of the communication symbols.}
The main goal of the adaptive abstractor is to determine a suitable state abstraction based on the learnt communication protocol, which reduces the size of the state space to be explored (the semantic problem), which in turn helps the agents achieve their goal, without much degradation in the policy performance (the effectiveness problem). Unlike \cite{StateAbstractionApprenticeship}, the proposed approach does not assume any prior knowledge of an expert policy to learn the abstraction. Moreover, the size of the abstracted space is not determined a priori. For the FoV-based navigation scenario studied in this work, a structured state of the environment is obtained by utilizing the Quad/Octree decomposition method \cite{QuadtreeHanan}. Simulation results show that the proposed method reaches a better policy, in terms of  achievable rewards in less number of training iterations, compared to the case when raw states are used for policy learning, or when a fixed state abstraction is used. 
Moreover, it is shown that the trained policy can generalize to cases when the local observations are uncertain about up to 60$\%$ of its local information due to noise, distortions, or obstructed views. 
Finally, it is shown that a communication protocol emerges during training which helps the agents learn better policies within fewer training iterations.

The rest of this paper is organized as follows. In Section \ref{sec:System-model}, the system model is introduced. The problem is formulated in Section \ref{sec:Problem-formulation}. In Section \ref{sec:ProposedAbstractor}, the proposed adaptive abstractor is presented followed by the training process. Finally, the simulation setup and results are discussed in Section \ref{sec:Numerical-results},  while  conclusions are drawn in Section \ref{sec:Conclusion}.

\section{System Model\label{sec:System-model}}

\begin{figure}
\centering \includegraphics[width=0.45\textwidth]{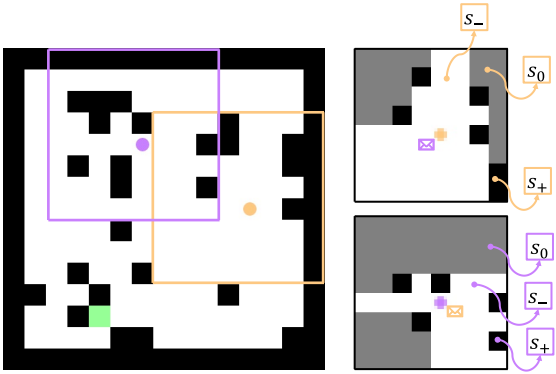}
\caption{A $15\times 15$ grid world (left) with obstacles (black tiles) and two agents (circles) having a partial observation of size $8\times8$. Agents cooperatively navigate to the destination (green tile) by only utilizing their partial observation and communicated symbols (right).}
\label{fig:1}
\end{figure}

As shown in Fig.~\ref{fig:1}, we consider $N$ autonomous agents navigating in a grid world of size $L\times L$. 
This grid world has one green goal and several obstacles with the density of $\varrho$.  
We assume that each agent $n$ knows its current location $l_n$ within the grid world. Additionally, each agent observes a partial view of the environment, of size $M\times M$, centered at its current location, where $M < L$, and $M$ is assumed to be a power of $2$. Moreover, agents cannot see through the obstacles. 
As a result, from any agent's perspective, the state at any location on the grid within the agent's FoV can be one of four cases:
\begin{itemize}
    \item Occupied ($s_{+}$): If an obstacle is sensed at that location.
    \item Unoccupied ($s_{-}$): If the location is obstacle-free.
    \item Destination ($s_{d}$): If the location is obstacle-free and contains the target destination.
    \item Unknown ($s_{0}$): If the location falls within the agent's blind spot, e.g., behind an obstacle.
\end{itemize}

Furthermore, to promote cooperation between agents, each agent is given access to a control channel to broadcast communication symbols to neighboring agents. In particular, each agent $n$ maintains a vocabulary $\mathcal{C}^n$ of size $C$, i.e., $|\mathcal{C}^n| = C$. These communication symbols $c\in\mathcal{C}^n$ have no predefined semantics, nor pre-specified usage rules \cite{Emergent_Comm-Lazaridou}. Instead, the agents should learn how to utilize these symbols while solving the navigation task, resulting in an emergence of a communication protocol.
Here, the cooperative objective of the agents is to navigate to the green goal location, by only utilizing their local partial observation and via communication leveraging the control channel.

\section{Problem Formulation \label{sec:Problem-formulation}}

The cooperative navigation task can be viewed as an MARL task with communication, which is modeled as a partially observable general-sum Markov game \cite{GroundEmergentComm,MarkovGames-LITTMAN}. Each agent obtains a partial observation of the global state at every time step. This partial observation includes the agent's current location, and the $M\times M$ partial view of the environment.\footnote{The union of all agents' partial observations forms the global state.} This observation, along with all symbols communicated on the control channel,  is used to learn a policy, to select a navigation action (up, down, left, right, and stay) and a communication symbol that maximizes the agent's reward. A reward of $1$ is received by each agent that completes the task. A deep neural network is used to parameterize the policy function.

Formally, a decentralized MARL can be described as a decentralized partially observable Markov decision process (Dec-POMDP), $\langle N, \mathcal{S}, \mathcal{A}, \mathcal{C}, \mathcal{O}, \mathcal{T}, R, \gamma \rangle$, where $N$ is the number of agents, $\mathcal{S}$ is the global state space, $\mathcal{A} = \{ \mathcal{A}^1, \dots, \mathcal{A}^N \}$, $\mathcal{C} = \{ \mathcal{C}^1, \dots, \mathcal{C}^N \}$, and $\mathcal{O} = \{ \mathcal{O}^1, \dots, \mathcal{O}^N \}$ are sets of actions, communications, and observation spaces, respectively.

At time-step $t$, each agent $n$ obtains a partial observation $o^{(n)}_t\in\mathcal{O}^n$ of the global state $s_t$, and a set of communicated symbols from the previous time step $c_{t-1} = \{ c_{t-1}^{(1)}, \dots, c_{t-1}^{(N)} \}$. The agent then selects an action $a_t^{(n)} \in \mathcal{A}^n$ and a communication symbol to broadcast $c_t^{(n)} \in \mathcal{C}^n$. Given the joint actions of all agents $a_t = \{ a_{t}^{(1)}, \dots, a_{t}^{(N)} \}$, the transition function $\mathcal{T} : \mathcal{S} \times \mathcal{A}^1 \times \dots \times \mathcal{A}^N \to \Delta (\mathcal{S})$ maps the current state $s_t$ and set of actions $a_t$ to $\Delta (\mathcal{S})$, a probability distribution over the next state $s_{t+1}$. 
Finally, each agent receives an individual reward $r_t^{(n)} \in R^{n}(s_t,a_{t}^{(n)})$ where $R^{n} : \mathcal{S} \times \mathcal{A}^n \to \mathbb{R} $.
A cooperative setting is considered for this task, where the objective of each agent is to maximize its expected return $\sum\limits_{t \in T} \gamma^t R^{n}(s_t, a_{t}^{(n)})$, 
%
%
%
%
for some finite time horizon $T$ and discount factor $\gamma$.

This objective is optimized using policy gradient \cite{GroundEmergentComm}. Here, any policy optimization algorithm can be used to optimize the policy network. However, directly using raw pixel observations as states poses a huge challenge for the learning problem. This is due to the unstructured nature of pixel data \cite{GroundEmergentComm}, as well as the huge observation space that needs to be explored to optimize the policy network \cite{MARLsurvey}. 

\begin{figure}
\centering \includegraphics[width=0.45\textwidth]{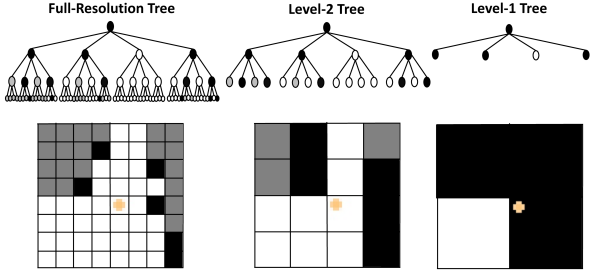}
\caption{Different levels of quadtree representation (top) with the corresponding observations (bottom).}
\label{fig:2}
\end{figure}

To overcome those pressing challenges, a structured-state abstraction is leveraged. State abstraction is a method for simplifying complex problems into simpler forms by compressing the state space into a structure-preserving manner  \cite{StateAbstractionApprenticeship,StructureAbstraction}. State abstraction could be carried out by eliminating unimportant states or combining states into groups to create a smaller state space \cite{StateAbstractionAPreview}. However, if abstraction compresses too much information, this would affect the agent's performance. Thus, care must be taken to identify the optimal state abstraction that balances between an appropriate degree of compression and acceptable representational power.
Representational power refers to the ability of the abstracted state to provide the agent with the necessary information to achieve its task. If the abstracted state doesn't have enough representational power (too much abstraction), then the agent will not be able to learn an effective policy. On the other hand, if the abstracted state has too high representational power, then the agent will have to learn a more complex policy. Fig.~\ref{fig:2} can be a good example of different representation levels.
Moreover, if the state space is structured, this  allows the agent to form concepts by forming equivalent classes, and use them for better generalization \cite{StructureAbstraction}.

In order to carry out state abstraction, the unstructured raw pixel observations can be transformed into a structured observation space by utilizing the concept of region quadtree decomposition \cite{QuadtreeHanan}\footnote{If 3D observation is available, Octree decomposition \cite{QuadtreeHanan} could be utilized without any changes to the proposed solution.}. Quadtree is a tree data structure used to efficiently store and represent two-dimensional space data, e.g.,  $M\times M$ partial view of the environment observed by each agent. A quadtree decomposition recursively decomposes a two-dimensional space into four equal sub-regions till all the locations within a sub-region have the same state or until reaching a maximum predefined resolution (tree-depth). Moreover, a full-resolution quadtree is defined as a quadtree that is decomposed until the finest granularity is reached.
 
A quadtree can be modeled as an unweighted graph $G=(\mathcal{V},\mathcal{E})$ where $\mathcal{V}$ is the set of nodes in the quadtree and $\mathcal{E}$ is the set of branches. Moreover, each node $v \in \mathcal{V}$ has a feature vector $X_v$ that consists of the node index as well as its state. 
Fig.~\ref{fig:2} shows different levels of quadtree representations of the agent's partial observation.

Abstracting the resultant quadtree structure can now be done by adaptively trimming branches of the quadtree structure. For  abstraction to be deemed useful and to enhance the agent's performance, abstraction should be a function of the agent's partial observation and the communication symbols received from other agents. In this paper, we propose a local adaptive abstractor at each agent that learns which quadtree branches to trim (abstracting the state space) depending on the current partial observation and  received communication symbols. Note that the communicated symbols will guide the abstraction process, depending on how they are interpreted by the agents.

\section{Proposed Adaptive Abstractor and Training Process \label{sec:ProposedAbstractor}}

\begin{figure*}
\centering \includegraphics[width=0.9\textwidth]{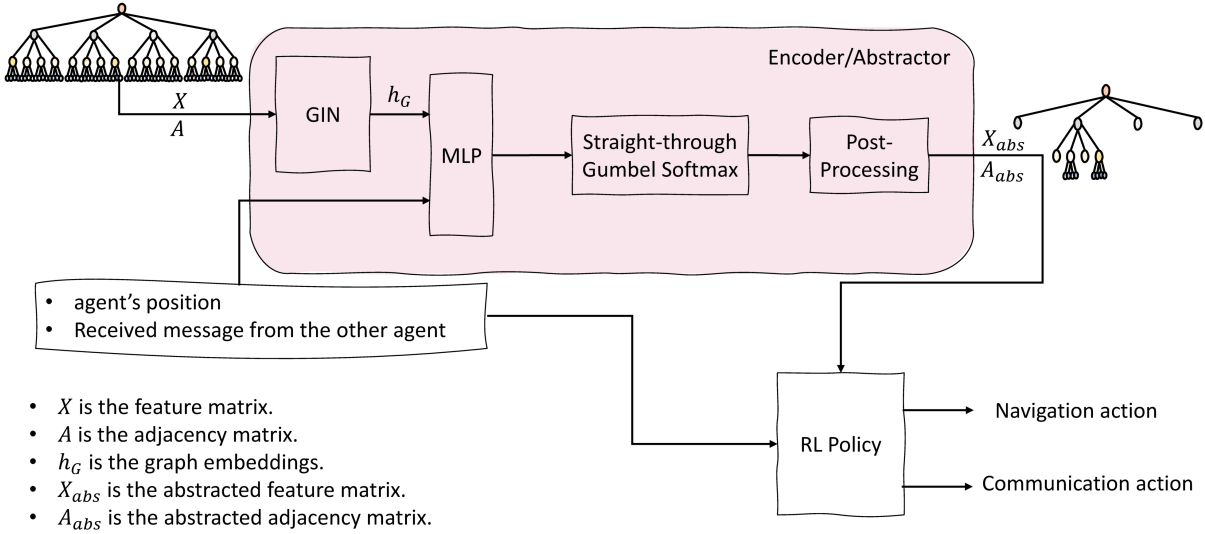}
\caption{Neural network architecture of the proposed abstractor at each agent.
}
\label{fig:NN_Abstractor}
\end{figure*}

In order to learn which quadtree branches to trim without affecting the agent's policy performance, we propose a neural network architecture for learning the abstractor.
Here, the abstractor neural network uses the full-resolution quadtree of the agent's $M\times M$ partial view of the environment, its current location, and the received communication symbols as input and  then outputs the trimmed quadtree structure. 
This output along with the communicated symbols is then used to learn a policy that maximizes the agent's reward. As mentioned earlier,  abstraction  reduces the size of the observation space to be explored during policy learning, without much degradation in the policy performance. The proposed neural network architecture of the abstractor is presented in Fig.~\ref{fig:NN_Abstractor}. The details of each component will be discussed in the following subsections.

\subsection{Graph Isomorphism Network}

\begin{figure}
\centering \includegraphics[width=0.45\textwidth]{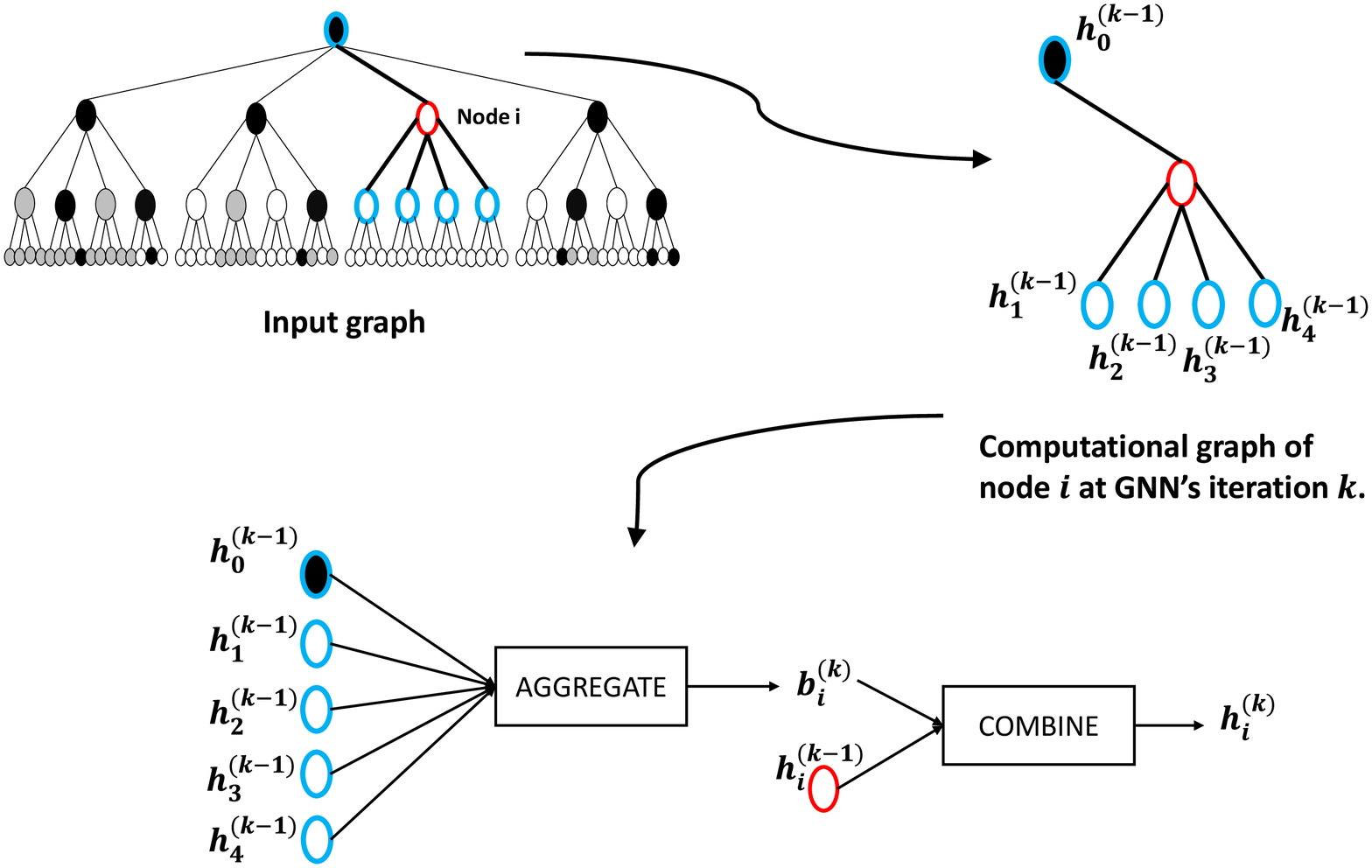}
\caption{An illustration of the $k^{\text{th}}$ iteration (or layer) of a general GNN model.}
\label{fig:GIN}
\end{figure}

Learning with graph structured data, such as quadtrees, demands an effective representation of this structure \cite{PowerfulGNN}. Representation learning of such graph structures is made possible by graph neural networks (GNNs). GNNs utilize the graph structure and node features to learn a representation (or an embedding) of the entire graph. GNNs follow a neighbourhood aggregation strategy where the representation of a node is iteratively updated by aggregating representations of its neighbours. After the $k$'s iteration, a node's representation captures structural information within its $k$-hop network neighbourhood. At the final iteration $K$, a readout function aggregates the representations of the nodes to obtain the entire graph's representation $h_G$. The $k$-th and final iteration $K$ of a GNN can be represented as follows:
\begin{align}\label{eq:Agg}      
    b_v^{(k)} & = \text{AGGREGATE}^{(k)} \left( \left\lbrace h_u^{(k-1)}  : u \in \mathcal{N}(v) \right\rbrace \right), \\  
    \label{eq:combine}h_v^{(k)}  & = \text{COMBINE}^{(k)} \left( h_v^{(k-1)}, b_v^{(k)} \right), \\
    \label{eq:readout}h_G        & = \text{READOUT} \left( \lbrace h_v^{(K)} \ \big\vert \ v \in G \rbrace \right)
\end{align}
where $h_v^{(k)}$ is the feature vector of node $v$ at the $k$-th iteration\footnote{$h_v^{(0)}$ is initialized to the node features $X_v$.}, and $\mathcal{N}(v)$ is the set of nodes adjacent to $v$. Fig.~\ref{fig:GIN} illustrates the GNN operations. The choice of $\text{AGGREGATE}^{(k)}$, $\text{COMBINE}^{(k)}$, and $\text{READOUT}$ functions in GNNs is crucial \cite{PowerfulGNN}.

For the GNN to have maximal representational power and disambiguate between different graph structures, i.e., by mapping them to different representations, the $\text{AGGREGATE}^{(k)}$, $\text{COMBINE}^{(k)}$, and $\text{READOUT}$ functions need to be injective \cite{PowerfulGNN}. As a result, graph isomorphism network (GIN) represents \eqref{eq:combine} and \eqref{eq:readout} as follows:
\begin{align}      
    h_v^{(k)}  & = \text{MLP}^{(k)} \left( h_v^{(k-1)} + \sum\nolimits_{u \in \mathcal{N}(v)} h_u^{(k-1)} \right), \\
    h_G        & = \text{CONCAT} \left( \sum\nolimits_{v \in G} h_v^{(k)} \ \big\vert  k = 0, 1, \ldots, K \right).
\end{align}
where $\text{MLP}$ is a multi-layer perceptron, and $\text{CONCAT}$ is the concatenation function.  

Due to its representational power that could distinguish between different structures by mapping them to different representations\footnote{The ability to map any two different structures to different embeddings entails solving the graph isomorphism problem. For more details, please refer to \cite{PowerfulGNN}.}, GIN is utilized in our proposed architecture. 
As a summary, the input of the GIN in our framework is the full-resolution tree represented as (i) the feature matrix $X$ which includes the feature vector of each node in the quadtree, and (ii) the adjacency matrix $A$ where an entry $a_{ij}$ is equal to $1$ only if there is a branch (or edge) between node $i$ and node $j$ in the quadtree structure. Otherwise it is equal to $0$. The output of the GIN is the entire graph's representation $h_G$.
 
\subsection{MLP and Straight-through Gumbel softmax}

After obtaining the graph representation $h_G$, it needs to be combined with the other information available at the agent, specifically, the agent's location and the received communication symbols from the previous time slot. This combining is done by concatenating all these information and passing them to an MLP. Note that, the internal nodes\footnote{Internal nodes are the nodes of the quadtree that are neither a root nor a leaf.} of the input tree-structure are the nodes that could be merged. As a result, the output of the MLP is set to be $2 \times |\mathcal{V^*}|$, where $\mathcal{V^*}$ is the set of internal nodes. The reason behind this is to compute a merge and non-merge score for each internal node and sample according to this score by utilizing straight-through Gumbel-softmax estimator (ST-GS) \cite{ST-GS}. ST-GS uses the $\text{argmax}$ operator over Gumbel random variables to generate a discrete random outcome in the forward pass and it maintains differentiability in the backward pass. The detailed information about the ST-GS estimator can be found in \cite{ST-GS}. The output of ST-GS in the forward pass is a binary vector yielding the merging decision of each internal node. 

\subsection{Post processing}
If the abstractor decides to merge a certain node, then all of its children should be discarded from the feature matrix $X$, and the adjacency matrix $A$. This will result in the abstracted (or trimmed) tree represented by the abstracted feature matrix $X_\text{abs}$ and the abstracted adjacency matrix $A_\text{abs}$.

\subsection{Decentralized training process \label{sec:Training}}

In this work, decentralized training is considered, where each agent is parameterized by an independent policy, which is a standard GRU policy with a linear layer. As mentioned earlier, any policy optimization algorithm could be used to optimize the policy network. 

Since no expert policy is assumed to be known, the training of the abstractor is carried out simultaneously while training and optimizing the policy network. The  training objective is to minimize the number of branches within the input tree, while still being able to complete the assigned task. Specifically, during training, the agent minimizes the following loss function $\mathbb{L}$:
\begin{equation}\label{eq:loss}
    \mathbb{L} = \mathbb{L}_\text{RL} + \lVert A_{\text{abs}} \rVert_\text{F},
\end{equation}
where $\mathbb{L}_\text{RL}$ is the RL loss dependent on the chosen policy optimization algorithm, and $\lVert A_{\text{abs}} \rVert_\text{F} = \sqrt{\sum_{i,j} {\lvert a_{ij} \rvert}^2}$ is the Frobenius norm of $A_\text{abs}$.
In this work, we use asynchronous advantage actor-critic (A3C) \cite{A3C} to optimize \eqref{eq:loss}, similar to \cite{GroundEmergentComm}. A3C is an actor-critic policy gradient algorithm that uses multiple CPU threads on a single machine. This would allow for a reduction in training time, as well as a stabilization in the learning process without the need for an experience replay \cite{A3C}. Since A3C is out of scope of this work, we refer interested readers to \cite{A3C} for more details on the training algorithm.

\section{Simulation setup and Analysis\label{sec:Numerical-results}}

In this section, we demonstrate the benefits of the proposed abstractor compared to several baselines. The evaluation is done in an environment called \texttt{FindGoal}, adopted from \cite{GroundEmergentComm}. A snapshot of this environment is shown in Fig.~\ref{fig:1}. In \texttt{FindGoal}, the task is to reach the green goal location. Each agent receives a reward of $1$ when they reach the goal. It should be noted that the states of the environment, e.g., locations of agents, goal and obstacles are randomized at every episode. Specifically, the map is a $15\times15$ grid world. At the beginning of each episode, a single goal tile and $25$ obstacle tiles are randomly placed on the map. Then, two agents are randomly placed on the remaining empty space. Each agent can only observe an $8\times8$ partial view of the map centered on the agent. Each agent has five actions: up, down, left, right, and stay. The maximum episode length allowed is $1024$ time steps. More simulation parameters are detailed in Table \ref{table:SimParam}.
For all simulations, the Adam optimizer \cite{AdamOptimizer} is used with a learning rate of $0.0001$ and parameters $\beta_1=0.9$, $\beta_2=0.999$, $\epsilon=10^{-8}$.

\subsection{Baselines}
We compare the proposed adaptive abstractor to four baselines detailed as follows:
\begin{itemize}
    \item \textbf{Full Resolution Image:} In this baseline, the local view of the agent, as in Fig.~\ref{fig:1}, is processed using a convolutional neural network (CNN)\footnote{The CNN architecture is as follows: 4 convolutional layers, where each layer has a kernel size of 3, stride of 2, padding of 1, and outputs 32 channels. ELU activation is applied to each convolutional layer. A 2D adaptive average pooling is applied over the output from convolutional layers. The final output has 32 channels, a height of 3, and a width of 3.}, and the output is passed to the RL policy, along with the agent's position and the received symbol from the other agent.
    \item \textbf{Full Resolution Tree:} In this baseline, the full resolution image is decomposed into the full-resolution quadtree, as shown in Fig.~\ref{fig:2}. The full-resolution quadtree, represented by its feature matrix and adjacency matrix, is processed by GNN without using the abstractor. The output is passed to the RL policy, along with the agent's position and the received symbol from the other agent.
    \item \textbf{Level-2 Tree:} Here, the full resolution image is decomposed into the level-2 quadtree, as shown in Fig.~\ref{fig:2}. This tree, represented by its feature matrix and adjacency matrix, is processed by GNN, and the output is passed to the RL policy, along with the agent's position and the received symbol from the other agent.
    \item \textbf{Level-1 Tree:} Here, the full resolution image is decomposed into the level-1 quadtree, as shown in Fig.~\ref{fig:2}. This tree, represented by its feature matrix and adjacency matrix, is processed by GNN, and the output is passed to the RL policy, along with the agent's position and the received symbol from the other agent. 
\end{itemize}
Level-1/level-2 tree can be seen as the output of a fixed abstractor that always abstract the quadtree to the first/second level in contrast to the proposed abstractor that adaptively trim branches of the quadtree depending on the system state.

\begin{table}[t]
\centering
\caption{Simulation parameters}\label{table:SimParam}
\begin{tabular}{c c c c c} 
 \cline{1-2}
 \cline{4-5}
 \textbf{Parameter} &
 \textbf{Value} &
 $\phantom{x}$&
 \textbf{Parameter} &
 \textbf{Value} \\ 
 \cline{1-2}
 \cline{4-5}
 $L$ & 15 && $\varrho$ & 0.15 \\ 
 $M$ & 8 && $C$ & 1024 \\
 $N$ & 2 && $\gamma$ & 0.99 \\
 $K$ & 5 && $T$ & 1024 \\
\cline{1-2}
 \cline{4-5}
\end{tabular}
\end{table}

\subsection{Training  performance evaluation}\label{subsec:training_performance}

\begin{figure}
\centering \includegraphics[width=0.45\textwidth]{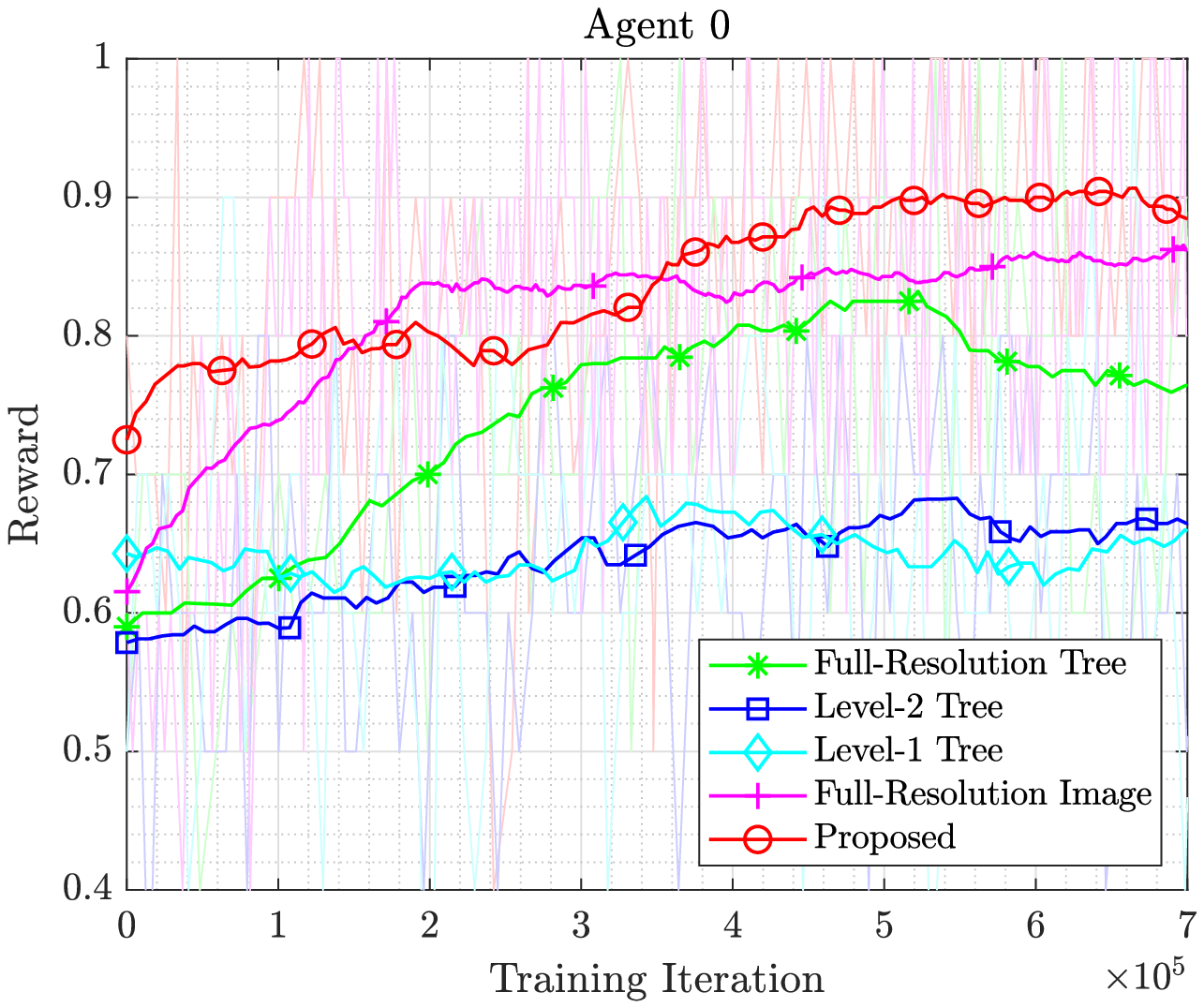}
\caption{The reward achieved by agent $0$ throughout training.}
\label{fig:Reward0}
\end{figure}

\begin{figure}
\centering \includegraphics[width=0.45\textwidth]{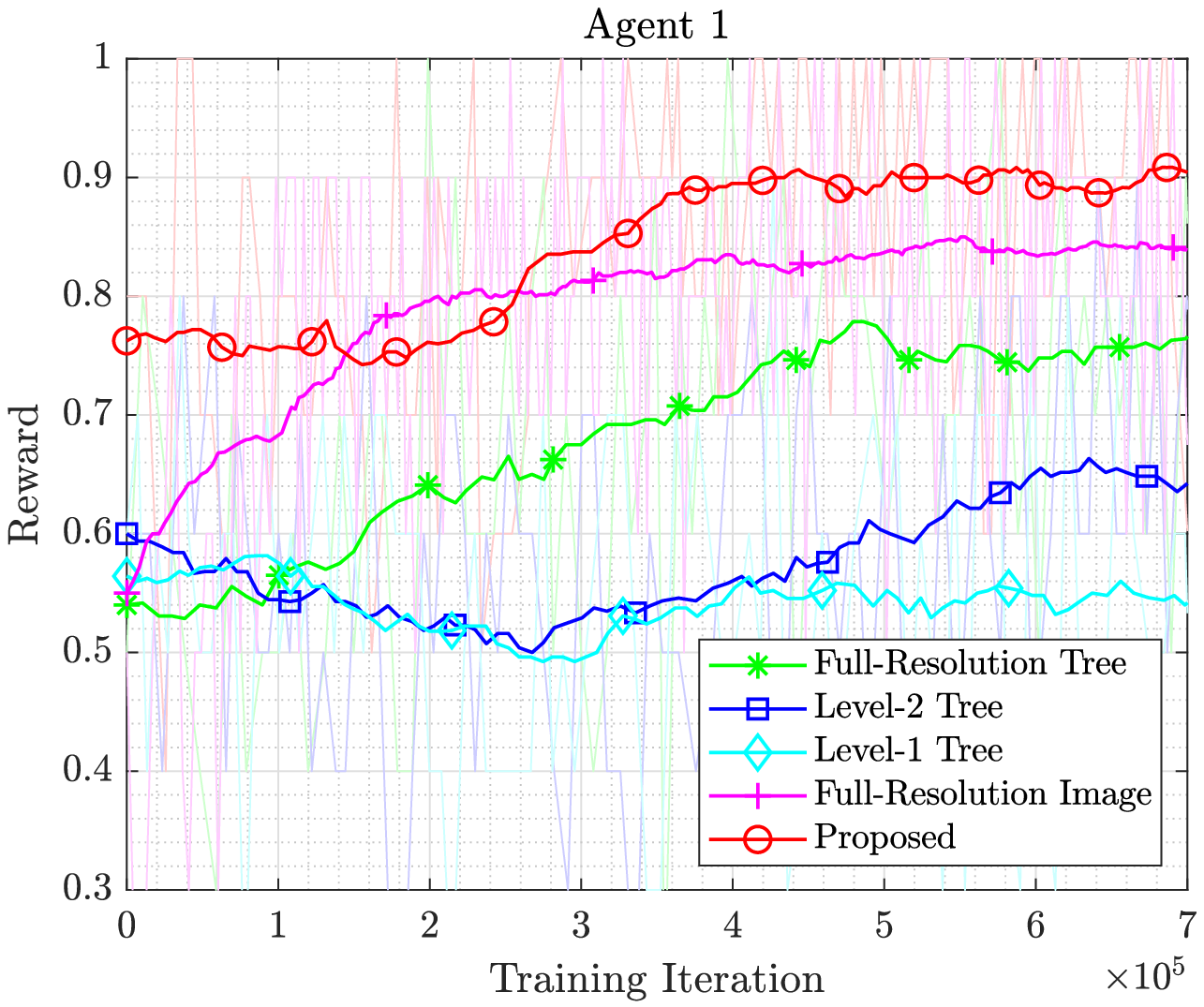}
\caption{The reward achieved by agent $1$ throughout training.}
\label{fig:Reward1}
\end{figure}

\begin{figure}
\centering \includegraphics[width=0.45\textwidth]{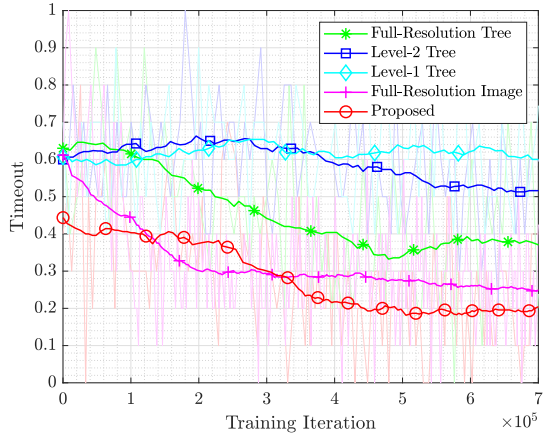}
\caption{Timeout during the evaluation episodes throughout training.}
\label{fig:Timeout}
\end{figure}

First of all, in Figs.~\ref{fig:Reward0}, \ref{fig:Reward1}, and \ref{fig:Timeout}, we evaluate the training speed and the performance of the trained policy using the proposed adaptive abstractor (\textbf{Proposed}), and  compare it with the four baselines outlined earlier. 

Figs.~\ref{fig:Reward0} and \ref{fig:Reward1} show the reward achieved by each agent throughout training. The faded background curves are the actual achieved reward, where each point over these curves is the average reward over $10$ evaluation episodes. 
These curves are smoothed, to gain insights about the training process, by a moving average over a window size of $200\,000$ training iterations (or episodes). The smoothed curves are represented by solid lines. We  observe that the baselines with fixed abstractions, e.g., \textbf{Level-1 Tree} and \textbf{Level-2 Tree}, fail to learn good policies during  training. This is due to the loss of useful information under  abstraction that neglects the current observation and the received communication symbols from the other agent. 
It can be noted that the adaptive abstraction (\textbf{Proposed}) method learns a policy that has better performance in terms of the achieved reward compared to \textbf{Full-Resolution Image} and \textbf{Full-Resolution Tree} baselines, although these baselines utilize all the available information without any abstraction. 
This can be attributed to the adaptive capability of the proposed abstractor that trims the input tree depending on the current observation and the received communication symbols, which yields a low-complexity yet informative tree structure.
Hence, the state space to be explored is reduced resulting in faster and better learning compared to the full-resolution baselines.

Fig.~\ref{fig:Timeout} shows the rate of failure of the agents during the evaluation episodes throughout the training process. Here, an episode is considered as a failure if any of the agents did not reach the goal location within the allowed maximum episode length.
The fraction of failures compared to total episodes is defined as the rate of failure, which is referred to as \textbf{Timeout} hereinafter. Similar to Figs. ~\ref{fig:Reward0} and \ref{fig:Reward1}, the background curves are the achieved reward over short intervals, where each point corresponds to the average reward over ten evaluation episodes. These curves are smoothed by the moving average over a window size of $200\,000$ training iteration (or episode) and represented by the solid lines. We notice the same trend in the performance as with the achieved reward by the agents, where the proposed adaptive abstractor outperforms the baselines and achieves a lower timeout after fewer training iterations. This is due to  the abstractor's capability of extracting low-complexity yet informative tree structures, as discussed under the results of Figs. ~\ref{fig:Reward0} and \ref{fig:Reward1}.


\subsection{Abstraction}

One of the compelling question we want to address is ``how much abstraction is achieved using the proposed adaptive abstractor?".
In order to measure the degree of abstraction, we resort to $\lVert A \rVert_\text{F} = \sqrt{\sum_{i,j} {\lvert a_{ij} \rvert}^2}$, which is directly proportional to the tree size.

\begin{figure}
\centering \includegraphics[width=0.45\textwidth]{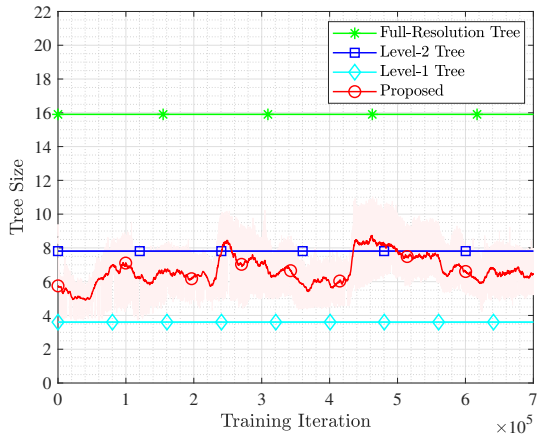}
\caption{Size of the tree during the training process.}
\label{fig:AbstractionLoss}
\end{figure}

The tree size of the proposed adaptive abstractor during the training process along with the tree sizes of the tree-based baselines: \textbf{Full-Resolution Tree}, \textbf{Level-1 Tree}, and \textbf{Level-2 Tree} are presented in Fig.~\ref{fig:AbstractionLoss}. 
Since the tree-based baselines can be considered as approaches with fixed abstraction, their tree sizes are represented by straight lines. Note that, the size of the tree achieved by the proposed method lies between \textbf{Level-1} and \textbf{Level-2} trees. Despite of that, the trained policy using the proposed abstractor outperforms that of the baselines, as shown in Section \ref{subsec:training_performance}. This can be attributed to the adaptive capability of the abstractor that trims the input tree as a function of the agent's local information and received message while preserving the useful information.

\begin{figure}
\centering \includegraphics[width=0.45\textwidth]{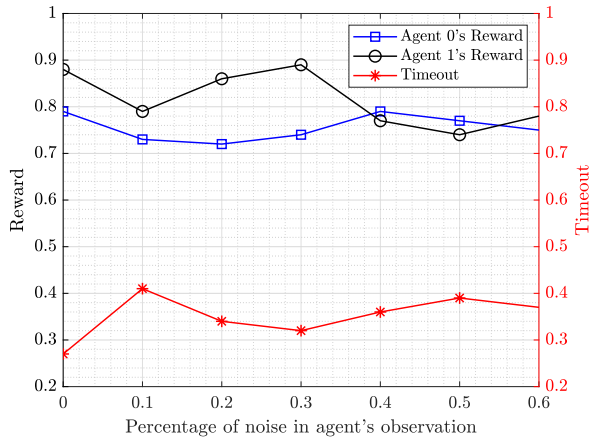}
\caption{Reward and timeout achieved by a trained policy over different values of noise probability levels $\alpha$.}
\label{fig:GeneralizationNoisyObs}
\end{figure}

\begin{figure}
\centering \includegraphics[width=0.45\textwidth]{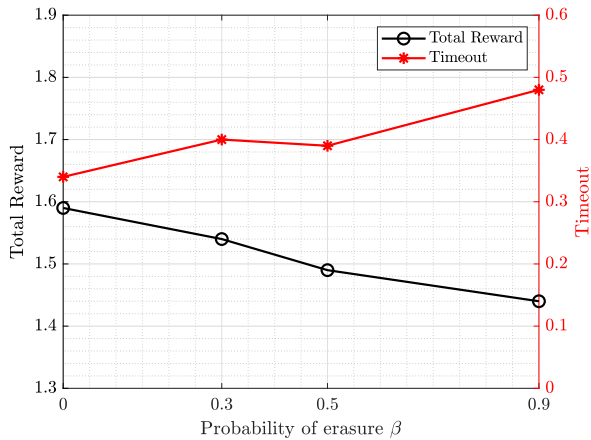}
\caption{Total reward and timeout achieved by a trained policy over different values of channel erasure probability $\beta$.}
\label{fig:GeneralizationErasureCh}
\end{figure}

\subsection{Generalization} 

In order to evaluate the generalization aspects of the trained policy, we adopt new situations where the local observation of each agent is distorted with noise.
The noisy observations are characterized by a noise level $\alpha$. Specifically, during evaluation, the occupied $s_+$ and unoccupied $s_-$ states of the locations within an agent's FoV are observed correctly by the agent with probability $1-\alpha$, and are seen as unknown states $s_0$ with probability $\alpha$. This scenario models the practical case where  sensors on a deployed agent are worn out or noisy, thereby affecting the agent's local observation and navigation task.

We run $100$ evaluation episodes utilizing the trained policies for different values of $\alpha$. Fig.~\ref{fig:GeneralizationNoisyObs} shows the achieved performance where each point is the average over $100$ evaluation episodes. Note that the policy is trained in noise-free setting, i.e., with $\alpha = 0$. We observe that the policy generalizes well with only a small degradation in performance, even when $\alpha=0.6$. This can be attributed to the structured input space, which allows the agent to see similarities and form concepts, enabling the agent to generalize better \cite{StructureAbstraction}.

Moreover, in order to assess the sensitivity of the trained policy to link level quality, we evaluate the trained agents in an environment with erasure channels. In such environment, the symbols transmitted over these channels can be erased with a certain probability $\beta$, and can be received correctly with probability $1-\beta$. \footnote{Although erasure channels may not be a perfect model for a communication scenario, they are a simple and effective way to model the effects of noise and interference on communication links and it is also a good model for the effects of severe path loss, where the signal is so weak that it is undetectable.} 

We run $100$ evaluation episodes utilizing the trained policies for different values of $\beta$. 
Fig.~\ref{fig:GeneralizationErasureCh} shows the achieved performance where each point is the average over the evaluation episodes. Note that the policy is trained with an ideal transmission conditions, i.e., with $\beta = 0$. We observe that the trained policy generalizes well, with an acceptable degradation in performance even when $\beta=0.5$.

\begin{figure}
\centering \includegraphics[width=0.45\textwidth]{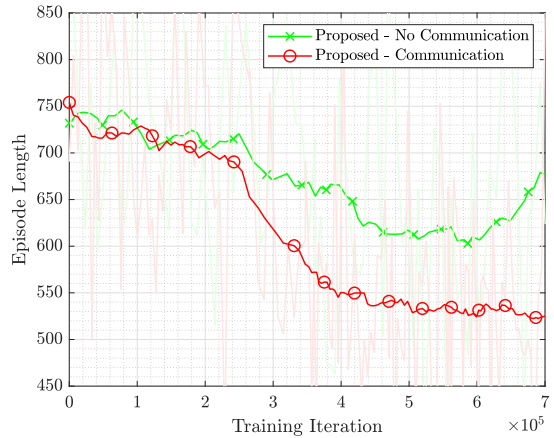}
\caption{Episode length throughout training using the proposed method with and without communication.}
\label{fig:EpisodeLengthCommVsNoComm}
\end{figure}

\begin{figure}
\centering \includegraphics[width=0.45\textwidth]{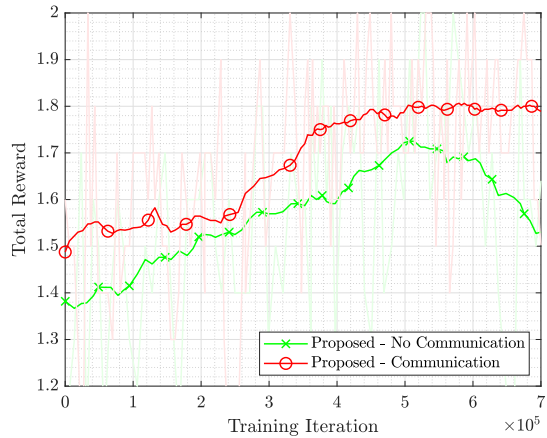}
\caption{Total reward achieved by the agents throughout training using the proposed method with and without communication.}
\label{fig:TotalRewardCommVsNoComm}
\end{figure}
\subsection{Emergent communication}

Finally, in order to detect whether communication has emerged at all during the training process, we  train the same scenario without a communication channel and compare the achieved performance of both cases, i.e., with and without communication \cite{PitfallsEC}. In Figs.~\ref{fig:EpisodeLengthCommVsNoComm} and \ref{fig:TotalRewardCommVsNoComm}, we analyze  the episode length and the total reward achieved during the training process by the proposed method, with and without communication. We can see that  when a communication channel exists among the agents, the performance is improved significantly.  This corroborates the fact that when agents communicate, they can help each other to complete the task much better than working independently. 
It should be noted that at the beginning of the training, the messages had no prior meaning while towards the end, a meaning has been emerged allowing both agents to cooperate well.


\section{Conclusion \label{sec:Conclusion}}

In this work, we have proposed a novel adaptive state abstraction for a cooperative FoV-based navigation scenario. In particular, we proposed a  neural network architecture to jointly learn an adaptive state space abstraction, along with a communication protocol among  agents to enhance coordination and cooperation. The main objective of the abstractor is to reduce the size of the state space to be explored during training, without much degradation in the policy performance. Simulations have shown that  the state space  can be significantly decreases and hence the complexity of RL policy training, without sacrificing the performance. Moreover, it was shown that the trained policy can generalize to cases when the local observations are uncertain about up to $60\%$ of its contents due to noise, distortions or obstructed views. Finally, it is shown that communication emerges during training which helps the agents to learn better policies within fewer training iterations. Future work will aim at analysing the properties of the emerged communication protocol among the agents, as well as generalization to more agents.

\bibliographystyle{IEEEtran}
\bibliography{IEEEabrv,Draft}

\begin{IEEEbiography}
[{\includegraphics[width=1in,height=1.25in,clip,keepaspectratio]{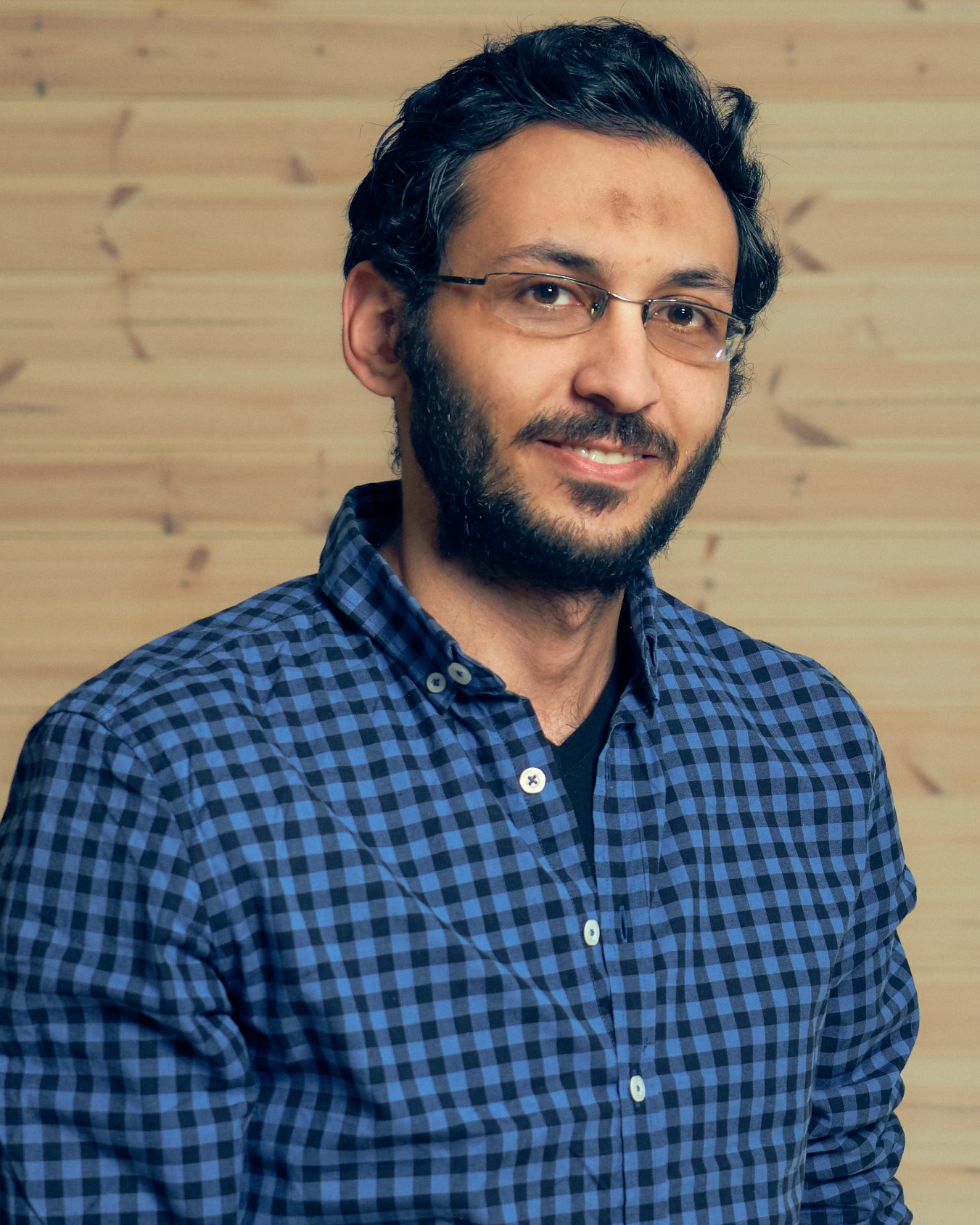}}]
{Mohamed K. Abdel-Aziz}
is with the Network systems and security research lab at Nokia Bell Labs since 2023. He received the B.Sc. degree (Hons.) in electronics and communications engineering from Alexandria university, Egypt, in 2013, and the M.Sc. degree in communication and information technology from Nile University, Egypt, in 2016. He is currently pursuing the Ph.D. degree with the University of Oulu, Finland under the Intelligent Connectivity and Networks/Systems Group (ICON) in the Centre for Wireless Communications (CWC). His research interests include ultra-reliable low-latency communications, vehicular networks, wireless artificial intelligence, and intent-based management.
\end{IEEEbiography}

\begin{IEEEbiography}
[{\includegraphics[width=1in,height=1.25in,clip,keepaspectratio]{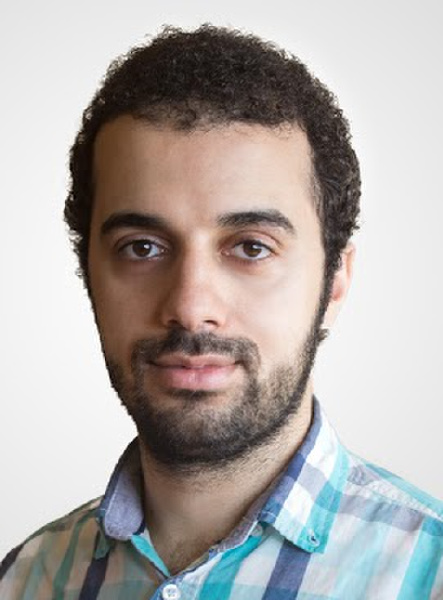}}]
{Mohammed S. Elbamby}
is with the Network Systems and Security Research Lab at Nokia Bell Labs since 2020. He received the doctoral degree (with distinction) from the University of Oulu, Finland, in 2019, and the master's degree from Cairo University, Egypt, in 2013, both in Communications Engineering. Previously, he held research positions at the University of Oulu, The University of Hong Kong, Cairo University, and the American University in Cairo. His research interests span network automation, context-awareness, application-network interaction, and machine learning for mobile networks. He received the Best Student Paper Award from the European Conference on Networks and Communications (EuCNC'2017).
\end{IEEEbiography}

\begin{IEEEbiography}
[{\includegraphics[width=1in,height=1.25in,clip,keepaspectratio]{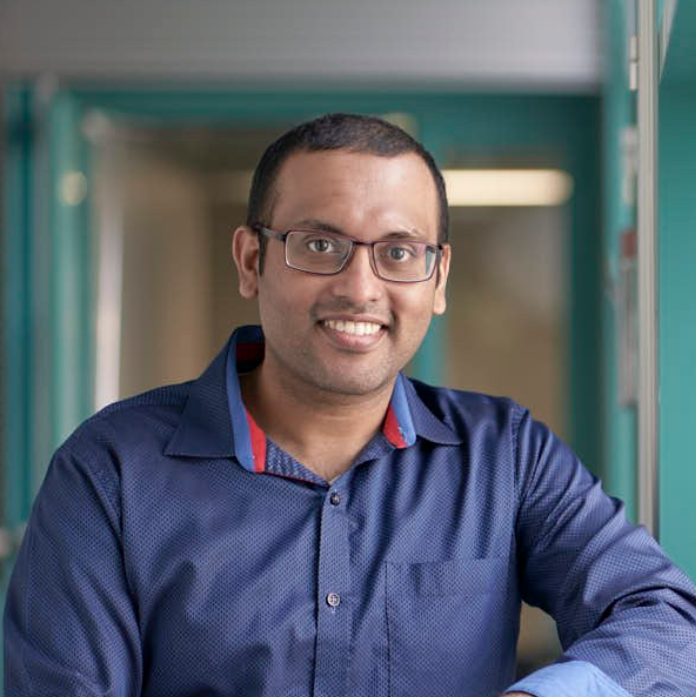}}]
{Sumudu Samarakoon}
received his B. Sc. Degree (Hons.) in Electronic and Telecommunication Engineering from the University of Moratuwa, Sri Lanka in 2009, the M. Eng. degree from the Asian Institute of Technology, Thailand in 2011, and Ph. D. degree in Communication Engineering from University of Oulu, Finland in 2017. Currently he is an Assistant Professor in Centre for Wireless Communications, University of Oulu, Finland and a member of the intelligent connectivity and networks/systems (ICON) group. His main research interests are in heterogeneous networks, small cells, radio resource management, reinforcement learning, and game theory. In 2016, he received the Best Paper Award at the European Wireless Conference and Excellence Awards for innovators and the outstanding doctoral student in the Radio Technology Unit, CWC, University of Oulu.
\end{IEEEbiography}

\begin{IEEEbiography}
[{\includegraphics[width=1in,height=1.25in,clip,trim=200 0 500 150,keepaspectratio]{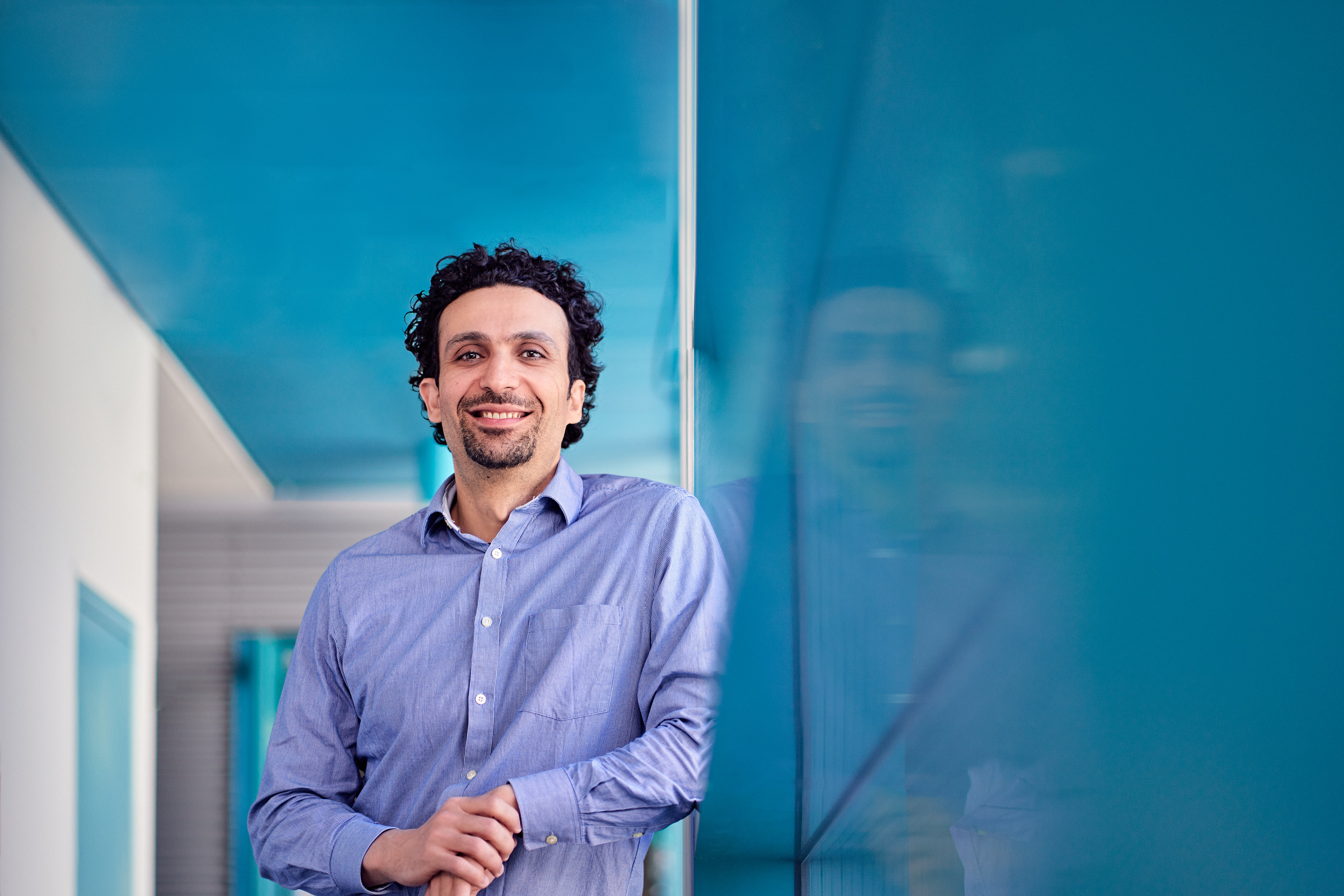}}]
{Mehdi Bennis}
is a Professor at the Centre for Wireless Communications, University of Oulu, Finland, IEEE Fellow and head of the intelligent connectivity and networks/systems group (ICON). His main research interests are in radio resource management, game theory and distributed AI in 5G/6G networks. He has published more than 300 research papers in international conferences, journals and book chapters. He has been the recipient of several prestigious awards including the 2015 Fred W. Ellersick Prize from the IEEE Communications Society, the 2016 Best Tutorial Prize from the IEEE Communications Society, the 2017 EURASIP Best paper Award for the Journal of Wireless Communications and Networks, the all-University of Oulu award for research, the 2019 IEEE ComSoc Radio Communications Committee Early Achievement Award and the 2020-2023 Clarviate Highly Cited Researcher by the Web of Science.
\end{IEEEbiography}

\end{document}